\documentclass[10pt,twocolumn,letterpaper]{article}

\usepackage{cvpr}              
\definecolor{cvprblue}{rgb}{0.21,0.49,0.74}
\usepackage[pagebackref,breaklinks,colorlinks,allcolors=cvprblue]{hyperref}

\usepackage{textcomp, gensymb}
\usepackage{booktabs}
\usepackage{multirow}
\usepackage{colortbl}
\usepackage{amssymb}
\usepackage[table]{xcolor}
\usepackage{overpic}
\usepackage{soul}
\usepackage{subcaption} 
\usepackage{amsmath}

\usepackage{rotating}

\usepackage[capitalize]{cleveref}
\crefname{section}{Sec.}{Secs.}
\Crefname{section}{Section}{Sections}
\Crefname{table}{Table}{Tables}
\crefname{table}{Tab.}{Tabs.}

\def\paperID{ 9252 } 
\def\confName{CVPR}
\def\confYear{2026}

\title{  STCast: Adaptive Boundary Alignment for Global and Regional \\ Weather Forecasting  }

\author{%
 Hao Chen$^{1}$ \;\;     
 Tao Han$^{1}$\;\; 
 Jie Zhang$^{1\ast}$\;\; 
 Song Guo$^{1\ast}$  \;\; 
 Lei Bai$^{2}$ \\  
 $^1$Hong Kong University of Science and Technology (HKUST)\;\;
  $^2$Shanghai AI Laboratory\\
  {\tt\small $^\ast$Corresponding author  }
 {\tt\small \{hchener, thanad\}@connect.ust.hk }
 {\tt\small \{songguo, csejzhang\}@ust.hk }
}

\begin{document}
\maketitle

\begin{abstract}
To gain finer regional forecasts, many works have explored the regional integration from the global atmosphere, \textit{e.g.}, by solving boundary equations in physics-based methods or cropping regions from global forecasts in data-driven methods. However, the effectiveness of these methods is often constrained by static and imprecise regional boundaries, resulting in poor generalization ability. To address this issue, we propose Spatial-Temporal Weather Forecasting (STCast), a novel AI-driven framework for adaptive regional boundary optimization and dynamic monthly forecast allocation. Specifically, our approach employs a Spatial-Aligned Attention (SAA) mechanism, which aligns global and regional spatial distributions to initialize boundaries and adaptively refines them based on attention-derived alignment patterns. Furthermore, we design a Temporal Mixture-of-Experts (TMoE) module, where atmospheric variables from distinct months are dynamically routed to specialized experts using a discrete Gaussian distribution, enhancing the model’s ability to capture temporal patterns. Beyond global and regional forecasting, STCast is evaluated on extreme event prediction and ensemble forecasting. Experimental results demonstrate consistent superiority over other methods across all four tasks. Code: \url{https://github.com/chenhao-zju/STCast}

\end{abstract}

\section{Introduction}
\label{sec:intro}

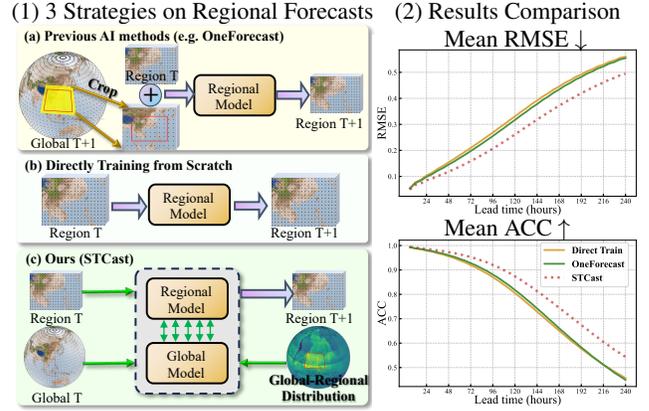
\begin{figure}[t]
    \centering
    \begin{overpic}[width=1.0\linewidth]{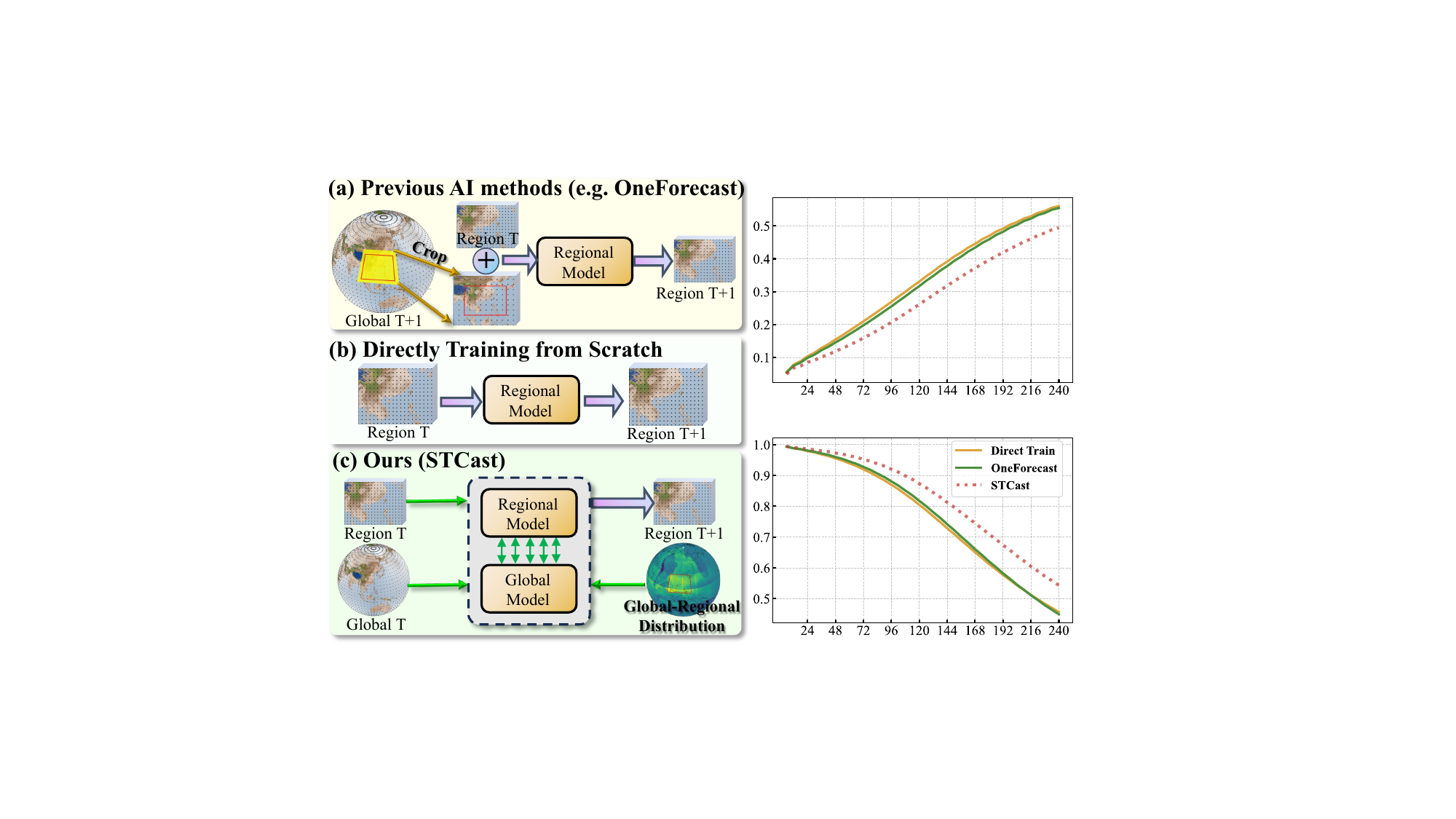} 
    \put(69.5,59.5){\footnotesize Mean RMSE $\downarrow$}
    \put(69.5,27.5){\footnotesize Mean ACC $\uparrow$}
    
    \put(3,64){\footnotesize (1) 3 Strategies on Regional Forecasts}
    \put(63.5,64){\footnotesize (2) Results Comparison}
    \end{overpic}
    \caption{(1) Illustration of 3 regional forecasting strategies: (a) Crop neighbor region from global forecasts and forecast with regional variables; (b) Directly training; (c) Forecast by densely connecting global-regional model with distribution. (2) Region forecasting comparison of 3 strategies. }
    \vspace{-5mm}
    \label{fig:sketch}
\end{figure}

\textbf{Why do we need global forecasts to support regional forecasting?} 
Achieving accurate, kilometre-scale regional weather forecasting is still a formidable scientific task with far-reaching socio-economic impact. Existing strategies typically fall into two paths: training a dedicated regional model or extracting the regional prediction from a global forecast. Traditional Numerical Weather Prediction (NWP)~\cite{nwp, lynch2008origins, kalnay2002atmospheric} methods solve partial differential equations (PDEs) at finer resolutions, but incur prohibitively high computational cost. Recent data-driven approaches~\cite{panguweather, fuxi, graphcast, aurora, subich2025fixing, weathergnn} significantly reduce this cost by neural networks. However, these models often rely on patch embeddings that downsample input variables, resulting in the loss of fine-grained local details. 
That is, training a global model at 1 km resolution (approximately 0.01\degree, or $19,980\times 39,960$) would be computationally infeasible. Conversely, restricting training to a high-resolution region neglects cross-regional dependencies that are critical for accurate forecasting. Thus, both direct training of high-resolution regional models and extracting them from high-resolution global forecasts are impractical. These limitations highlight the need for hybrid frameworks that couple low-resolution global forecasts with high-resolution regional forecasting in a computationally efficient manner.

\textbf{Why must adaptive boundaries cover the global area?} 
While hybrid global–regional frameworks are gaining attention, existing coupling strategies, whether based on traditional NWP~\cite{lundquist2010immersed, mani2012analysis} or AI models~\cite{oneforecast, mlwp, meps}, typically define regional boundaries using only adjacent areas.
This local perspective contradicts the well-established Atmosphere–Ocean–Land–Biosphere Coupling Theory~\cite{manabe1969climate, zhang2018ocean}, which posits that any points in the regional atmosphere are influenced by the entire Earth system. For example, Siberian cold surges can trigger East-Asian cold waves, and surface heating over the Tibetan Plateau can simultaneously alter East Asian monsoons and North American jet stream~\cite{wu2023integrated}. \textit{Thus, the true boundary for a region is not its neighbors, but the entire Earth.}

To address these challenges, we introduce \textbf{STCast}, a Spatial–Temporal Forecasting framework that explicitly models the evolving global–regional correlations within Earth system. Unlike prior methods that restrict the boundaries to neighboring regions, STCast initializes the global-regional distributions using spatial-aligned attention (SAA) and continuously refines them during training. Beyond spatial boundaries modeling, STCast further captures temporal variability by routing monthly atmospheric inputs to specialized experts via Temporal Mixture-of-Experts (TMoE), using a discrete Gaussian distribution. Together, SAA and TMoE enable STCast to deliver accurate and generalizable regional forecasts by incorporating both global spatial influences and fine-grained temporal patterns.


\textbf{Spatial-Aligned Attention (SAA)} incorporates a learnable global-regional distribution into linear cross-attention, enabling adaptive aggregation of global atmospheric information for regional forecasting. To couple the global and regional variables, SAA employs two key mechanisms: (1) a Great Circle distance metric to quantify spatial separation from the target region, and (2) an exponential distance-decay function to initialize the learnable global-regional distribution, ensuring weaker influence from distant regions. This prior modulates the attention weights by element-wise multiplication and is further refined during training. As a result, the global–regional correlation evolves dynamically, aligning spatial dependencies with physical intuition throughout the optimization process.
\textbf{Temporal Mixture-of-Experts (TMoE)} enhances the standard MoE framework by integrating a month-specific Gaussian prior to guide expert routing. It operates through three key mechanisms: (1) Learning a Gaussian distribution for each month to represent its temporal characteristics; (2) Modulating expert routing weights with this distribution, ensuring that weights decay with increasing temporal distance from an expert; (3) Enabling multi-expert activation to enhance routing diversity. This design facilitates dynamic input-to-expert assignment while preserving temporal specialization and improving generalization across time.

\textbf{Regional Forecasting Experiments.} As illustrated in \cref{fig:sketch}.(1), we compare three regional weather forecasting strategies, including previous AI methods (\cref{fig:sketch}a), directly training on the target region (\cref{fig:sketch}b), and our proposed STCast (\cref{fig:sketch}c). Unlike existing approaches that statically concatenate adjacent areas to the target region, STCast establishes a learnable global-regional distribution to adaptively aggregate low-resolution global forecasts into high-resolution regions. Quantitative results in \cref{fig:sketch}.(2) demonstrate that STCast achieves the best performance across all variables in terms of both Mean RMSE and ACC, outperforming Direct Train and OneForecast. 
These results validate the effectiveness of our dynamic, Earth-aware boundary mechanism over static neighbor-based coupling.
  
 

In conclusion, the contributions of this work include:
\begin{itemize}
    \item We propose an AI-based method to extract adaptive regional boundary from our Spatial-Aligned Attention (SAA) module. The approach is initialized with global-regional distribution and optimized during training.
    
    
    \item We introduce the Spatial-Temporal Forecasting Framework (STCast) for weather forecasting, featuring a novel Temporal Mixture-of-Experts (TMoE) architecture. This component dynamically allocates forecasting tasks across different months to specialized expert models, enhancing temporal adaptability.
    

    \item Extensive experiments across four critical weather forecasting tasks, including low-resolution global forecasts, high-resolution regional forecasts, typhoon track prediction, and ensemble forecasting, demonstrate that STCast achieves state-of-the-art performance, significantly outperforming existing methods.
    
\end{itemize}

\section{Related work}
\label{sec:related}

\begin{figure*}[t]
  \centering
   \includegraphics[width=0.98\linewidth]{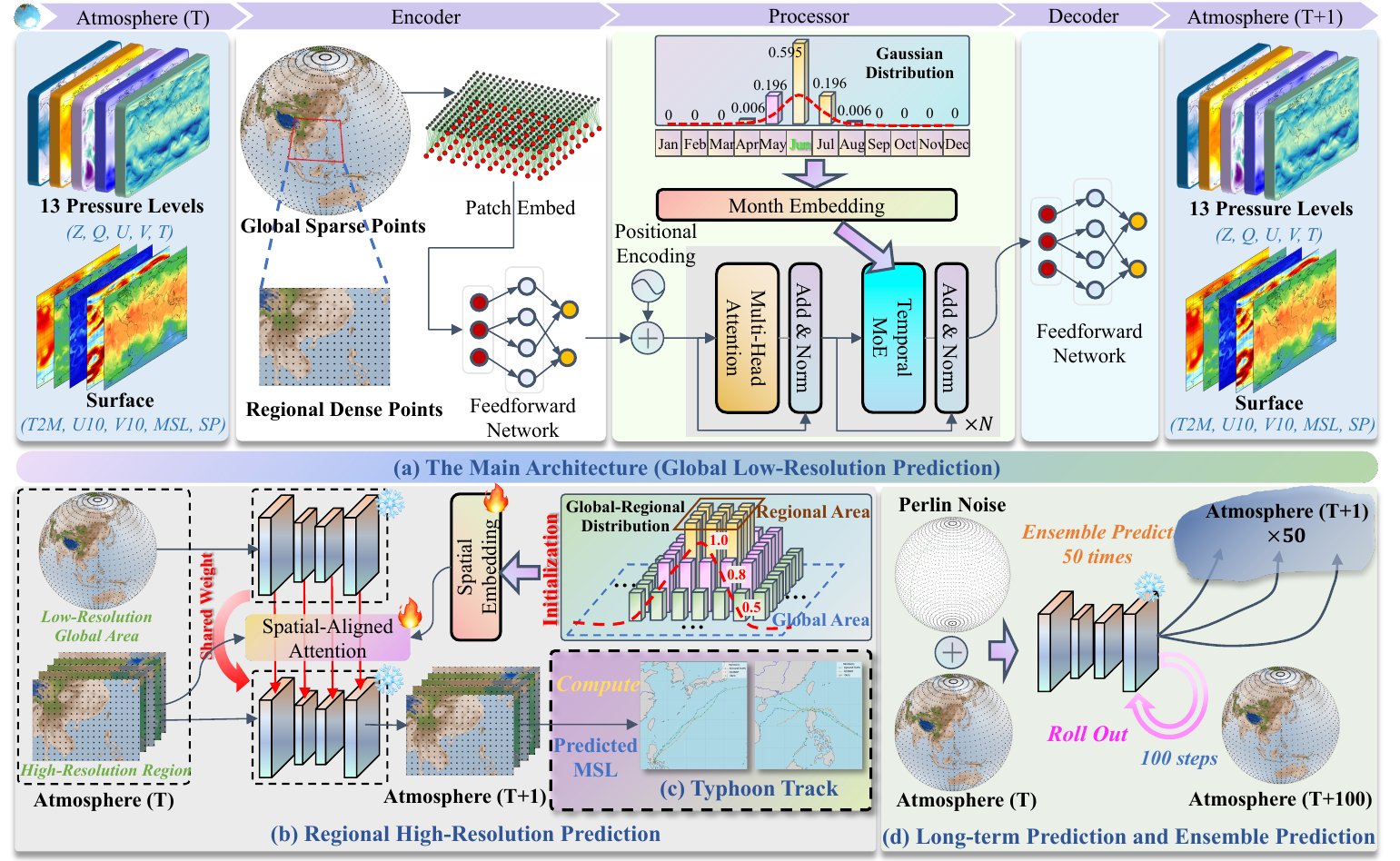}
   \vspace{-0.3cm}
   \caption{Illustration of our method. (a) The overall structure of low-resolution global weather forecasting, which includes input Atmospheric variables, an Encoder, a Processor, a Decoder, and output Atmospheric variables; (b) The high-resolution regional weather forecasting structure with Spatial-Aligned Attention (SAA) module; (c) The typhoon track prediction structure with predicted high-resolution MSL; and (d) The long-term weather forecasting and ensemble weather forecasting.
   }
   \label{fig:framework}
   \vspace{-12pt}
\end{figure*}

\subsection{ Global-Regional Weather Coupling } 



Accurate global–regional coupling remains a core challenge in regional forecasting due to the difficulty of boundary specification. NWP models address this by solving PDEs under prescribed boundary, \textit{e.g.}, sponge layers~\cite{mani2012analysis}, Dirichlet~\cite{hidayatullah2019topic}, and Neumann conditions~\cite{sabathier2023boundary}. In contrast, AI approaches~\cite{oneforecast, schmude2024prithvi, mlwp, meps} bypass boundary equations by concatenating a fixed neighborhood in global area to region, resulting in static and local coupling.

The recent work, OneForecast~\cite{oneforecast}, tackles the same 4 tasks as ours by concatenating neighboring low-resolution global forecasts with high-resolution regional variables. In contrast, our method replaces static concatenation with a transformer-based framework that adaptively models global–regional correlations, guided by a learned prior. This enables dynamic boundary refinement and long-range dependency modeling beyond local neighborhoods.

\subsection{ Data-Driven Weather Forecasting }

Prior to deep learning, NWP prevailed, producing forecasts by solving PDEs on high-resolution global grids~\cite{nwp, lynch2008origins, kalnay2002atmospheric}. These methods deliver physically analysis and rigorously validated forecasts~\cite{molteni1996ecmwf, ritchie1995implementation}, but their computation at inference remains prohibitively high, especially at fine scales.

The emergence of large-scale atmospheric reanalyses has catalysed a shift toward data-driven forecasting. Early exemplars, FourCastNet~\cite{fourcastnet} and Pangu-Weather~\cite{panguweather}, employ Fourier Neural Operators (FNO)~\cite{fno} and 3D Swin Transformers~\cite{liu2021swin}, respectively, to approximate atmospheric evolution. Subsequent research has bifurcated into two streams: (i) neural operators, such as KNO~\cite{koopmanlab} and SFNO~\cite{sfno}, that directly learn the temporal evolution operator; and (ii) neural networks, including FengWu~\cite{fengwu}, FengWu-ghr~\cite{han2024fengwu}, Graphcast~\cite{graphcast}, FuXi~\cite{fuxi}, GenCast~\cite{gencast}, and Stormer~\cite{stormer}, which leverage inductive biases tailored to atmosphere. All achieve competitive performance. 

Inspired by the success of MoE in LLMs~\cite{moe}, recent studies have begun to integrate MoE into weather forecasting. VAMoE~\cite{vamoe} extends MoE to incremental weather forecasting; EWMoE~\cite{ewmoe} augments FourCastNet with MoE layers. In contrast, we propose TMoE that explicitly partitions inputs by month and dynamically routes them to specialized temporal experts. This structure enables TMoE to capture inter-month variability and intra-month correlation.

\textbf{Additional Related Works} in time-series field~\cite{simvp, nmo, pastnet, stcwf, STTN, histgnn, pkdsttn} are provided in Appendix.

\section{Methodology} \label{sec:method}




We give a unified framework for 4 tasks: low-resolution global prediction, high-resolution regional prediction, typhoon track forecasting, and ensemble forecasting. ~\cref{subsec:problem} formally defines each task. We then introduce~\cref{subsec:saa} that fuses global and regional atmospheric variables via a learnable global-regional distribution, and~\cref{subsec:tmoe} that allocates monthly data to specialized experts.


\subsection{Overview} \label{subsec:problem}

STCast is composed of three components: the Encoder, Processor, and Decoder. The Processor employs an alternating strategy that integrates window-based attention with self-attention. This hybrid design allows the model to capture both local and global dependencies within the input distribution. \textbf{More details are provided in the Appendix.}

This framework is applied to the weather forecasting task, where the model \(\Phi\) predicts future atmospheric states \(\mathbf{X}^{t+1}\) based on historical inputs \(\mathbf{X}^{t}\). Specifically, \(\mathbf{X}^{t+1}=\Phi(\mathbf{X}^{t})\), where \(\mathbf{X}^{t}\) includes upper-air variables \(\mathbf{P}^{t}\in\mathbb{R}^{H\times W\times 13\times N}\) across 13 pressure levels and surface variables \(\mathbf{S}^{t}\in\mathbb{R}^{H\times W\times M}\), with N and M denoting the number of variables per pressure and surface level, respectively.

As illustrated in \cref{fig:framework}, STCast unifies four key subtasks to address diverse forecasting challenges: global deterministic forecasting \(\Phi_g\), high-resolution regional forecasting \(\Phi_r\), typhoon track prediction \(\Phi_{tc}\), and ensemble forecasting \(\Phi_{ens}\). To enhance temporal modeling, we introduce the Temporal Mixture-of-Experts (TMoE) and integrate Flash-Attention~\cite{flashattention} with MoE~\cite{moe}, as shown in \cref{fig:framework}(a). Considering the significant seasonal variability in atmospheric dynamics, TMoE assigns monthly forecasts to specialized experts. For global forecasting, this strategy is expressed as \(\mathbf{X}^{t+1}_{g}=\Phi_g(\mathbf{X}^{t}_{g})\), where \(\mathbf{X}^{t}_{g}\) denotes the global input variables. In regional forecasting, we adopt a global–regional coupling approach as depicted in \cref{fig:framework}(b). The Spatial-Aligned Attention (SAA) module fuses global \(\mathbf{X}^{t}_{g}\) and regional inputs \(\mathbf{X}^{t}_{r}\) to produce high-resolution predictions: \(\mathbf{X}^{t+1}_{r}=\Phi_r(\mathbf{X}^{t}_{r},\mathbf{X}^{t}_{g})\). The predicted mean sea level pressure (MSL) is subsequently used to infer typhoon tracks illustrated in \cref{fig:framework}(c). Beyond deterministic tasks, we also evaluate probabilistic forecasting for long-range and ensemble scenarios as shown in \cref{fig:framework}(d). To this end, we inject Perlin noise \(\mathbf{N}_g\) into the initial global state \(\mathbf{X}^{t}_{g}\) and execute ensemble simulation for n times. The ensemble mean \(\mathbf{X}^{t+1}_{g} = \frac{1}{n}\sum{i=1}^{n}\Phi_{ens}(\mathbf{X}^{t}_{g},\mathbf{N}_g)\) is used to provide probabilistic predictions.

The principal contributions of this work are the TMoE and SAA modules, detailed in the following subsections.

\begin{figure}[t]
  \centering
   \includegraphics[width=0.95\linewidth]{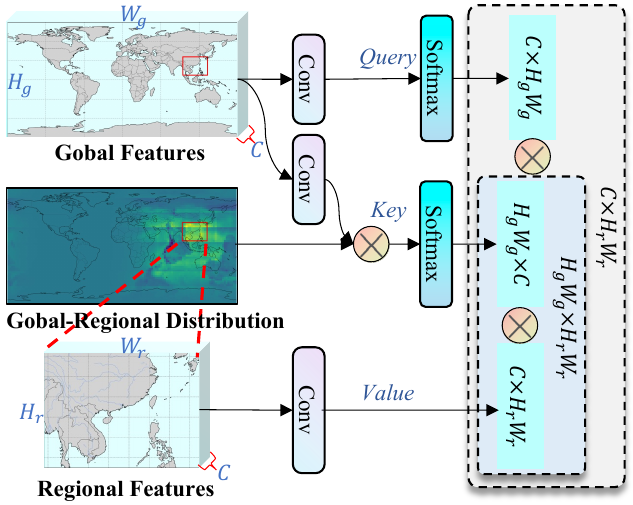}
   \vspace{-0.05cm}
   \caption{Illustration of Spatial-Aligned Attention.}
   \label{fig:saa}
   \vspace{-0.3cm}
\end{figure}

\begin{figure}[t]
  \centering
   \includegraphics[width=0.94\linewidth]{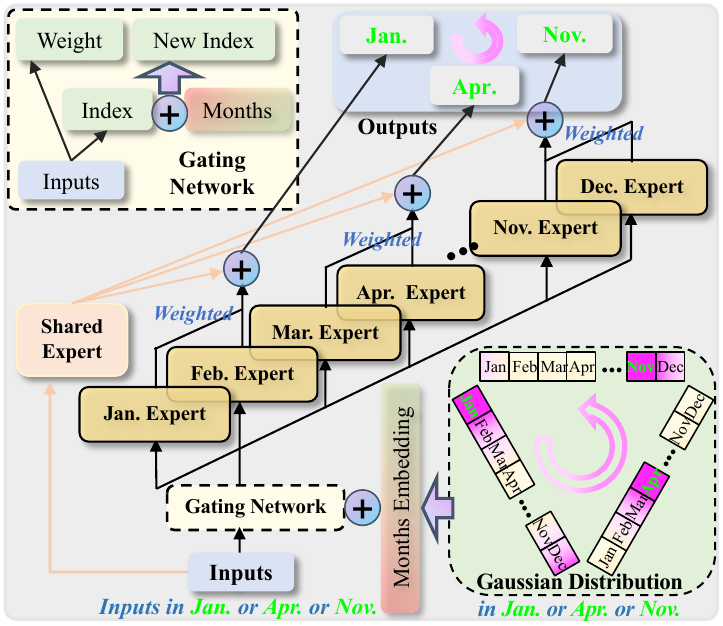}
   \vspace{-0.05cm}
   \caption{Illustration of Temporal Mixture of Experts.}
   \label{fig:tmoe}
   \vspace{-0.3cm}
\end{figure}

\subsection{Spatial-Aligned Attention} 
\label{subsec:saa}


As shown in \cref{fig:saa}, the Spatial-Aligned Attention (SAA) module employs global features $\textbf{X}^t_g \in \mathbb{R}^{H_g \times W_g \times C}$ as Query and Key, while utilizing regional features $\textbf{X}^t_r \in \mathbb{R}^{H_r \times W_r \times C}$ as Value. Unlike previous approaches that rely on static boundaries, our SAA module dynamically couples global and regional features through linear cross-attention at each block. This innovative design learns global-regional distribution from attention maps while maintaining computational efficiency through linear attention mechanisms, which effectively reduce the processing overhead.



To achieve precise quantification of spatial relationships, SAA calculates Great Circle distance between each global point and target region. Compared to Manhattan and Euclidean distance, the Great Circle distance provides a more accurate measure of spatial relationships on the Earth's surface. This distance metric is defined as:
\begin{align}
    d(\phi, \lambda) &= \max\left( 
    d_{\phi} - \frac{1}{2} H_r, 
    d_{\lambda} - \frac{1}{2} W_r
    \right), \\
    d_\phi &= R \cdot \arccos\left( \sin\phi \sin C_\phi + \cos\phi \cos C_\phi \right), \\
    d_\lambda &= R \cdot \arccos\left( \cos(\lambda - C_\lambda) \right).
\end{align}
where $\phi, \lambda$ represent the latitude and longitude coordinates, $(C_\phi, C_\lambda)$ denote the latitude and longitude of the target region's center, $(H_r, W_r)$ define the region's height and width, and $d_\phi, d_\lambda$ indicate the distances in the latitude and longitude directions, respectively. This efficient formulation introduces negligible overhead.


Next, the global–regional prior is derived from an exponential distance-decay function that monotonically reduces correlation as distance increases. The function is:
\begin{equation}
    f(\phi, \lambda) = 
        \begin{cases} 
        1.0 & ,d(\phi, \lambda) \leq 0 \\
        \exp\left(-\alpha \cdot \left[d(\phi, \lambda)\right]^2\right) & ,d(\phi, \lambda) > 0 
        \end{cases} ,
\end{equation}
where $\text{exp}$ and $\alpha$ denote the base of the natural logarithm and the decay factor, respectively. 


Regarding the cross-attention mechanism, SAA incorporates an efficient attention~\cite{efficientattn} to compute global-regional correlations. This design reduces computational complexity from $O(n^2)$ to $O(n)$. The process is formulated as:
\begin{equation}
    E(Q, K, V) = \text{Softmax}(Q) \left( \frac{\text{Softmax}(f(\phi, \lambda) \cdot K)^T V}{\sqrt{d_k}} \right),
\end{equation}
where $d_k$ denotes the embedding dimension of Key.

SAA establishes an optimal distribution by computing the Hadamard product between the initial global-regional distribution and the attention map. This trainable prior distribution serves a dual purpose: it guides the optimization process while being progressively refined, capturing the spatial relationships and learning latent correlations between global and regional atmospheric patterns.

\subsection{Temporal Mixture-of-Experts} 
\label{subsec:tmoe}

Acknowledging the discrepancy of atmospheric variables across different months, the Temporal Mixture-of-Experts (TMoE) framework treats forecasting for each month as relatively independent tasks and organizes these tasks using the Mixture-of-Experts (MoE). To assign training tasks for different months to specialized experts, TMoE employs a rotating discrete Gaussian distribution that directs the experts in training atmospheric variables across various months. The peak of the Gaussian distribution is rotated to correspond with the month of the input variables. The discrete Gaussian distribution is defined as follows:
\begin{equation}
    f(x)=\frac{1}{\sigma \sqrt{2 \pi}} \exp \left(-\frac{(x-\mu)^{2}}{2 \sigma^{2}}\right),
\end{equation}
where $\mu$ and $\sigma$ are the mean and variance of the distribution, and $x$ is a discrete series, $x\in\text{[1,2,3,...,12]}$, denoting 12 months in one year. To fit the atmospheric dataset, those two hyper-parameters are set to learnable during training.


Following the discrete Gaussian series, we perform rotational alignment of the month series to correspond with input variables. This alignment ensures a monotonic decrease in activation probability as temporal distance from the target month increases. Through this mechanism, input variables become distinguishable by their month. The aligned month series is subsequently encoded into continuous embedding representations via a MLP. These temporal embeddings serve as latent features that inform and optimize the expert selection process within TMoE.


In TMoE, the gating network first derives a weight vector and an index tensor from the input variables $\textbf{X}^{t}$. Month-specific information is incorporated by concatenating this index with the 12-dimensional month embedding $\textbf{M} \in \mathbb{R}^{12 \times 1}$. The resulting feature is then fed into a softmax layer that selects the Top-K experts. These experts are subsequently activated to model the conditional distribution of inputs for current month. The entire procedure is formulated as:
\begin{equation}
\textbf{I} = \text{Softmax}(\text{Conv}(\textbf{X}^t) + \textbf{M}),
\end{equation}
where $\textbf{I}$ denotes index, which selects Top-K experts.

Compared to prior MoE methods that employ implicit expert allocation strategies with auxiliary losses, TMoE introduces an explicit month embedding mechanism to assign input variables to specialized experts with limited computation. This explicit guidance more effectively prevents MoE homogenization during training. 




\section{Experiments}
\label{sec:experiment}

\begin{table*}[t]
    \caption{Performance of Ours with 4 baselines on Global Weather Forecasting. A small RMSE (normalized, $\downarrow$) and a bigger ACC (denormalized, $\uparrow$) indicate better performance. The best results are in \textbf{bold}, and the second best are with \underline{underline}.
}
    \vspace{-0.3cm}
    \centering
     \makebox[\textwidth]{
\resizebox{\linewidth}{!}{
            \renewcommand{\multirowsetup}{\centering}
            \begin{tabular}{l|cc|cc|cc|cc|cc}
                \toprule
                \multirow{3}{*}{Model} &  \multicolumn{2}{c}{6-hour} & \multicolumn{2}{c}{1-day} & \multicolumn{2}{c}{4-day} & \multicolumn{2}{c}{7-day} & \multicolumn{2}{c}{10-day}   \\
                \cmidrule(lr){2-11}
               & RMSE$\downarrow$ & ACC$\uparrow$ & RMSE$\downarrow$ & ACC$\uparrow$ & RMSE$\downarrow$ & ACC$\uparrow$ & RMSE$\downarrow$ & ACC$\uparrow$ & RMSE$\downarrow$ & ACC$\uparrow$ \\

                \midrule
                Pangu-weather\cite{panguweather} &0.0826&0.9876 &0.1571&0.9581&0.3380&0.8167&0.5092&0.5738&0.6215&0.3542     \\
                Graphcast\cite{graphcast} & 0.0626  & 0.9928  & 0.1304  & 0.9705   & 0.2861  & 0.8705  & 0.4597  & 0.6692  & 0.6009   & 0.4275    \\
                Fuxi\cite{fuxi} &0.0987 & 0.9820 &0.1708&0.9511  &0.4128&0.7379& 0.5972&0.4446&  0.6981&0.2391  \\
                OneForecast\cite{oneforecast} &\textbf{0.0549} &\underline{0.9943} &\underline{0.1231}    &\underline{0.9737}     &\underline{0.2732} &\underline{0.8825} & \underline{0.4468} &\underline{0.6888} &\underline{0.5918} &\underline{0.4457}  \\
                \midrule
                
                
                Ours & \underline{0.0617} & \textbf{0.9956} & \textbf{0.1197} & \textbf{0.9740} & \textbf{0.2578} & \textbf{0.8927} & \textbf{0.4348} & \textbf{0.7019} & \textbf{0.5763} & \textbf{0.4715}  \\

                \bottomrule
            \end{tabular}
    }}
    \vspace{-3mm}
    \label{tab:global}
\end{table*}

\begin{figure*}[t]
  \centering
   \includegraphics[width=1.0\linewidth]{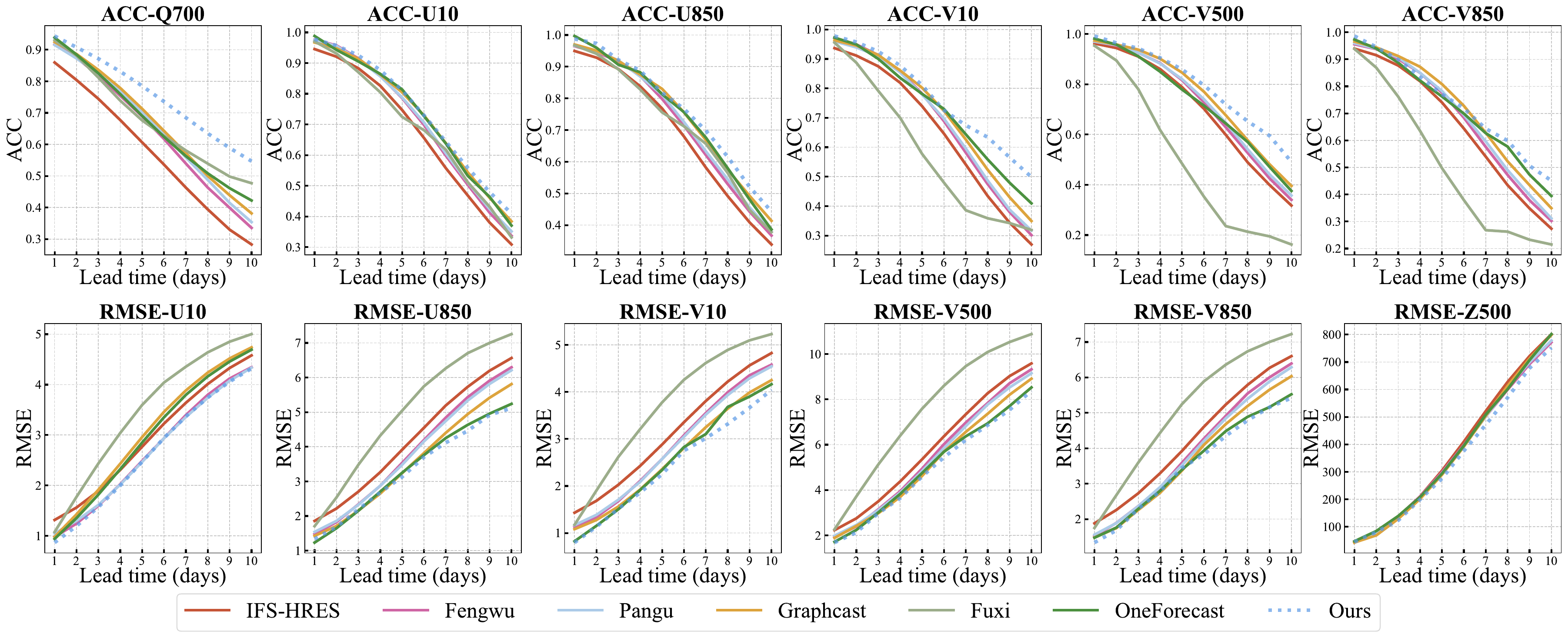}
   \vspace{-0.8cm}
   \caption{ Comparison of our method with 6 competitors on denormalized RMSE $\downarrow$ and ACC $\uparrow$ in Global Weather Forecasting. } 
   \label{fig:global}
   \vspace{-0.4cm}
\end{figure*}



\subsection{Dataset and Implementation Details} 

In this work, we conduct experiments on a popular weather dataset, \textit{i.e.}, ERA5~\cite{era5}, provided by the ECMWF~\cite{molteni1996ecmwf}. ERA5 dataset is a reanalysis atmospheric dataset, consisting of the atmospheric variables from 1979 to the present day with a 0.25\degree spatial resolution with $721 \times 1440$. The global model is trained on 40-year dataset from 1979 to 2019 with 70 variables, and the regional model is trained in the same period with 5 surface variables on Eastern Asia. 

We apply the AdamW optimizer with 0.0002 learning rate to model training. In both global and regional forecasts, we train 100 epochs and set batch size to 16. Our model is trained with 16 NVIDIA Tesla A100 GPUs. 

\textbf{Statement.} More details about Dataset and Implementation Details, Numerical Comparison and Computation Metric of typhoon track prediction, computation comparisons, additional results, additional visualization, and additional ablation studies are provided in Appendix.



\begin{table*}[t]
    
    \small
    \vspace{-2.1mm}
    \caption{ Ablation Studies on two tasks with normalized mean RMSE and denormalized mean ACC. }
    \vspace{-3mm}
    \centering
     \makebox[\textwidth]{
\resizebox{0.98\linewidth}{!}{
            \renewcommand{\multirowsetup}{\centering}
            \begin{tabular}{l|cc|cc|cc|cc|cc}
                \toprule
                \multirow{3}{*}{ Ablation Studies }  &  \multicolumn{2}{c}{6-hour} & \multicolumn{2}{c}{1-day} & \multicolumn{2}{c}{4-day} & \multicolumn{2}{c}{7-day} & \multicolumn{2}{c}{10-day}   \\
                \cmidrule(lr){2-11}
               & RMSE$\downarrow$ & ACC$\uparrow$ & RMSE$\downarrow$ & ACC$\uparrow$ & RMSE$\downarrow$ & ACC$\uparrow$ & RMSE$\downarrow$ & ACC$\uparrow$ & RMSE$\downarrow$ & ACC$\uparrow$ \\

                \midrule
                    & \multicolumn{10}{c}{  \textbf{High-resolution Regional Forecasts}  } \\
                \midrule

                \pmb{w/o} SAA & 0.0767  & 0.9762  & 0.1802  & 0.8675  & 0.4001   & 0.4127 & 0.5297  & 0.4229   & 0.7610    & 0.2541    \\
                
                \pmb{w/o} Global-Regional Distribution & 0.0694  &0.9805 &0.1631 &0.8864  &0.3794 &0.6718 & 0.5082 &0.4566 & 0.7192  &0.3286    \\
                
                \pmb{w} SAA & \textbf{0.0478}  & \textbf{0.9946}  & \textbf{0.0845}  & \textbf{0.9854}  & \textbf{0.2051}   & \textbf{0.9203}   & \textbf{0.3699}  & \textbf{0.7442}   & \textbf{0.4921}  &  \textbf{0.5433}    \\
                
                \midrule
                    & \multicolumn{10}{c}{ \textbf{Low-resolution Global Forecasts}  } \\
                \midrule

                \pmb{w/o} TMoE &0.0751 &0.9915 &0.1451 &0.9714 &0.3249 &0.8201 &  0.5109   & 0.5412    &  0.6426   & 0.3184   \\
                
                \pmb{w/o} Month Embedding  & 0.0744 & 0.9928 & 0.1346 & 0.9764 & 0.2865 & 0.8180 & 0.4631 & 0.5941    & 0.6049 & 0.3559  \\
                
                \pmb{w} TMoE & \textbf{0.0617} & \textbf{0.9956} & \textbf{0.1197} & \textbf{0.9740} & \textbf{0.2578} & \textbf{0.8927} & \textbf{0.4348} & \textbf{0.7019} & \textbf{0.5763} & \textbf{0.4715}  \\

                \bottomrule
            \end{tabular}
    }}
    
    \label{tab:ablation}
\end{table*}

\begin{figure*}[t]
    \centering
    \begin{overpic}[width=\linewidth]{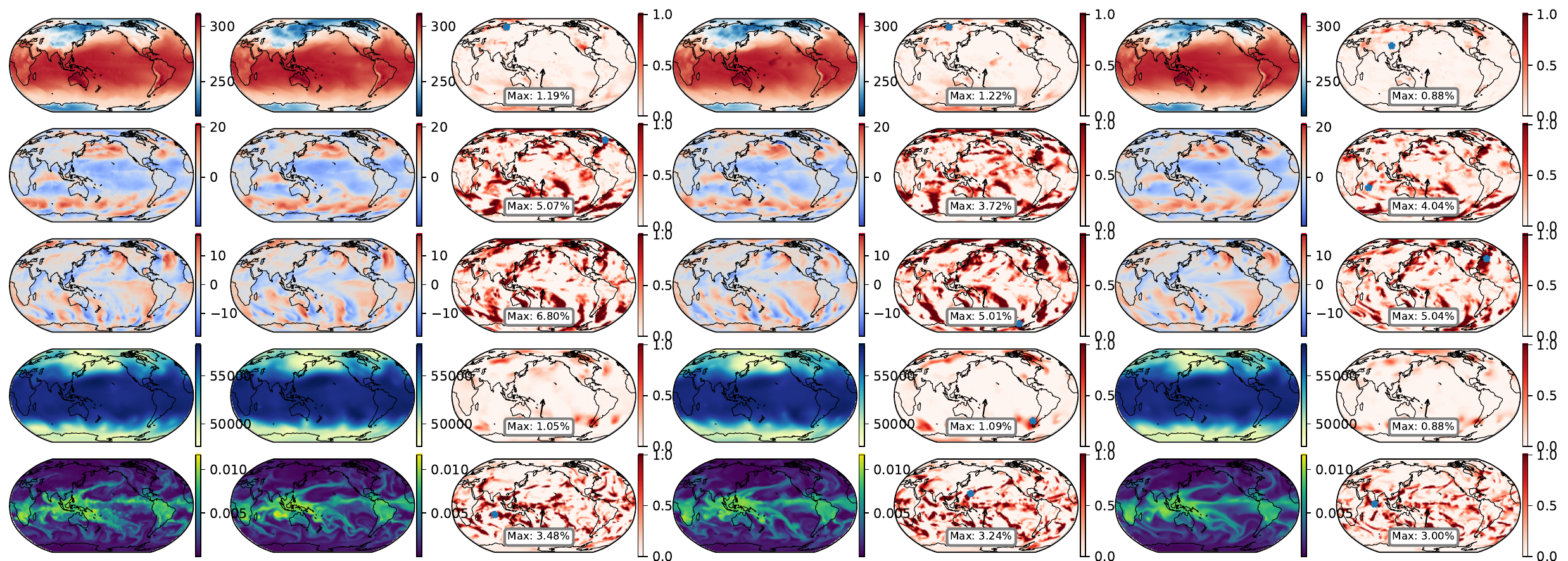} 


    \put(5.5,35){\scriptsize GT}
    \put(17.5,35){\scriptsize Graphcast}
    \put(29.5,35){\scriptsize Error@Graphcast}
    \put(45,35){\scriptsize OneForecast}
    \put(57,35){\scriptsize Error@OneForecast}
    \put(75.5,35){\scriptsize Ours}
    \put(87.5,35){\scriptsize Error@Ours}

    \put(-1, 1.5){\scriptsize \rotatebox{90}{ Q700 }}
    \put(-1, 9){\scriptsize \rotatebox{90}{ Z500 }}
    \put(-1, 16){\scriptsize \rotatebox{90}{ V10 }}
    \put(-1, 23){\scriptsize \rotatebox{90}{ U10 }}
    \put(-1, 30){\scriptsize \rotatebox{90}{ T2M }}
    
    \end{overpic}
    \vspace{-0.7cm}
    \caption{Visualization of 10-day global weather prediction on 5 variables among Graphcast, OneForecast, and Ours. }
    \vspace{-0.2cm}
    \label{fig:vis}
\end{figure*}

\begin{figure}[t]
    \centering
    \begin{overpic}[width=\linewidth]{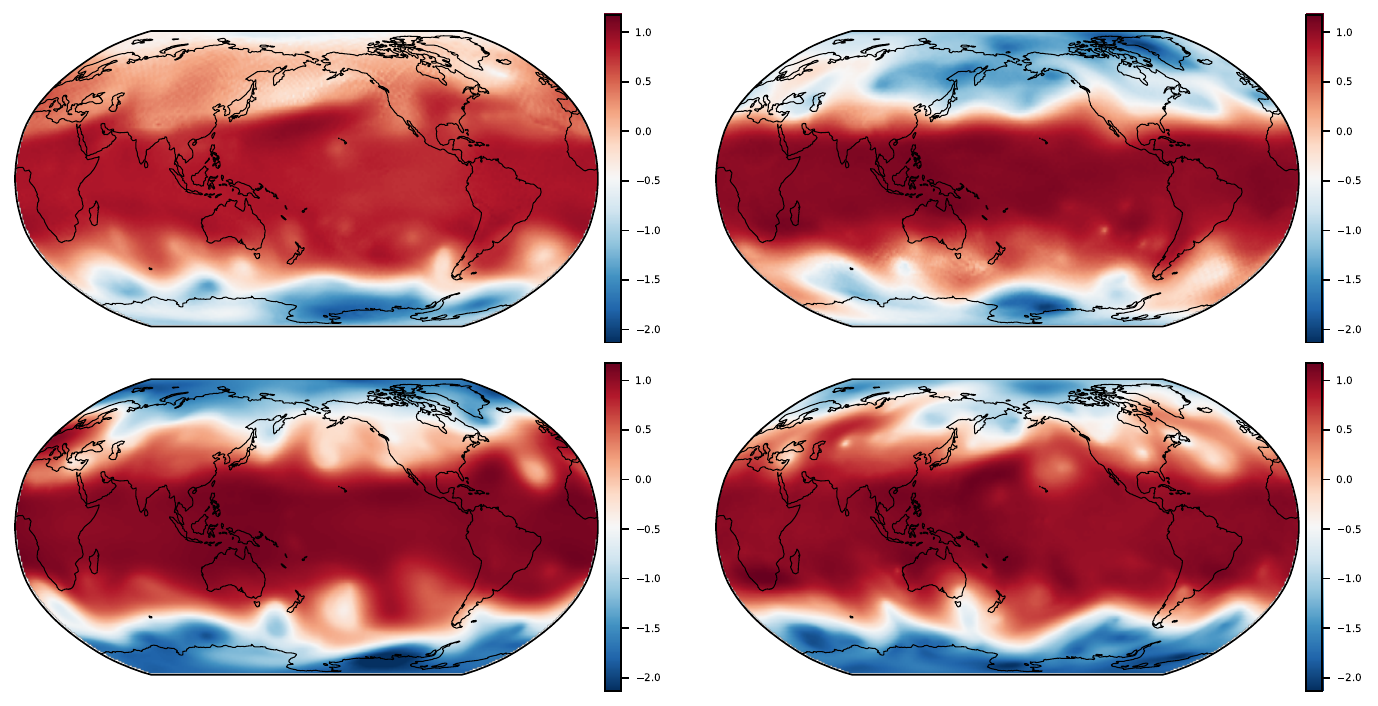} 
    \put(8.2, 49.6){\small Graphcast@Z500 }
    \put(57, 49.6){\small OneForecast@Z500 }
    \put(12.3, 24.3){\small Ours@Z500 }
    \put(64.3, 24.3){\small GT@Z500 }
    \end{overpic}
    \caption{Visualization of 100-day prediction of Z500 among Graphcast, OneForecast, and Ours. }
    \vspace{-5mm}
    \label{fig:long}
\end{figure}

\begin{figure*}[t]
    \centering
    \begin{overpic}[width=\linewidth]{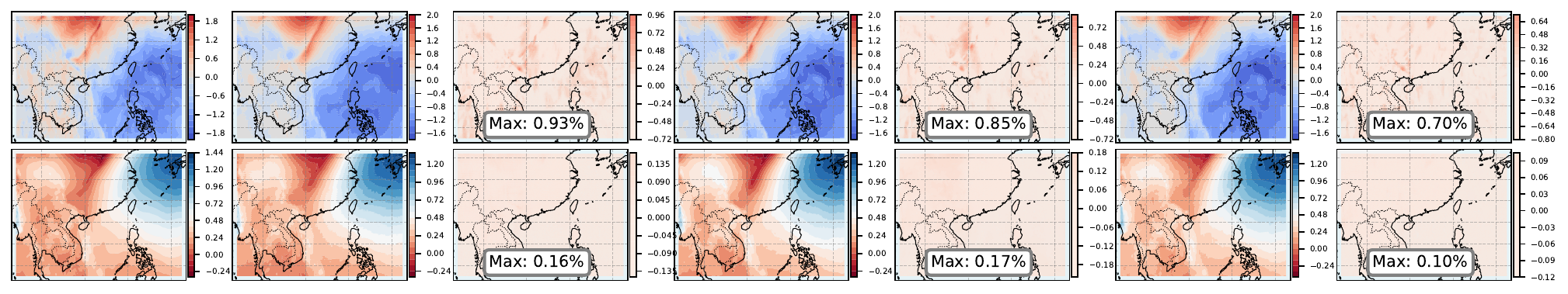} 

    \put(5.5,18.3){\scriptsize GT}
    \put(17,18.3){\scriptsize Direct Train}
    \put(29,18.3){\scriptsize Error@Direct Train}
    \put(45,18.3){\scriptsize OneForecast}
    \put(57,18.3){\scriptsize Error@OneForecast}
    \put(75.5,18.3){\scriptsize Ours}
    \put(87.5,18.3){\scriptsize Error@Ours}

    \put(-0.5, 3){\scriptsize \rotatebox{90}{ MSL }}
    \put(-0.5, 12){\scriptsize \rotatebox{90}{ U10 }}
    
    \end{overpic}
    \vspace{-8mm}
    \caption{Visualization of 6-hour regional weather prediction on MSL and U10 among Direct Training, OneForecast, and Ours. }
    \vspace{-5mm}
    \label{fig:region}
\end{figure*}
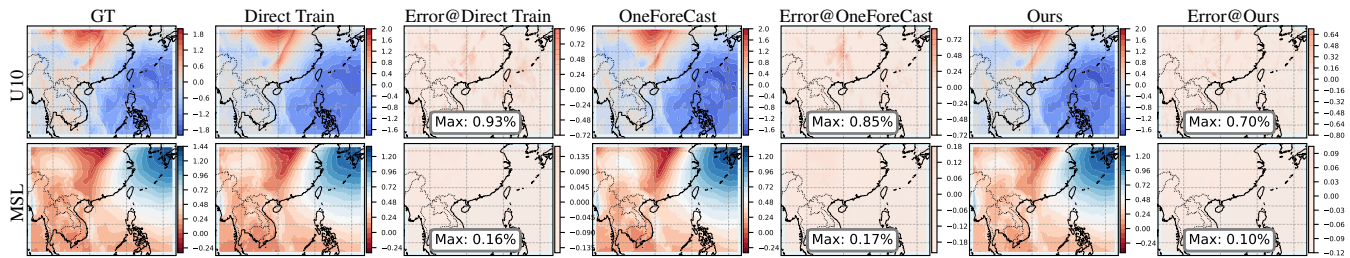

\begin{figure}[t] 
    \centering
    \begin{minipage}{\columnwidth} 
        \centering
        

        \begin{subfigure}{0.5\textwidth}
            \centering
            \captionsetup{justification=centering}
            \includegraphics[width=1.0\linewidth]{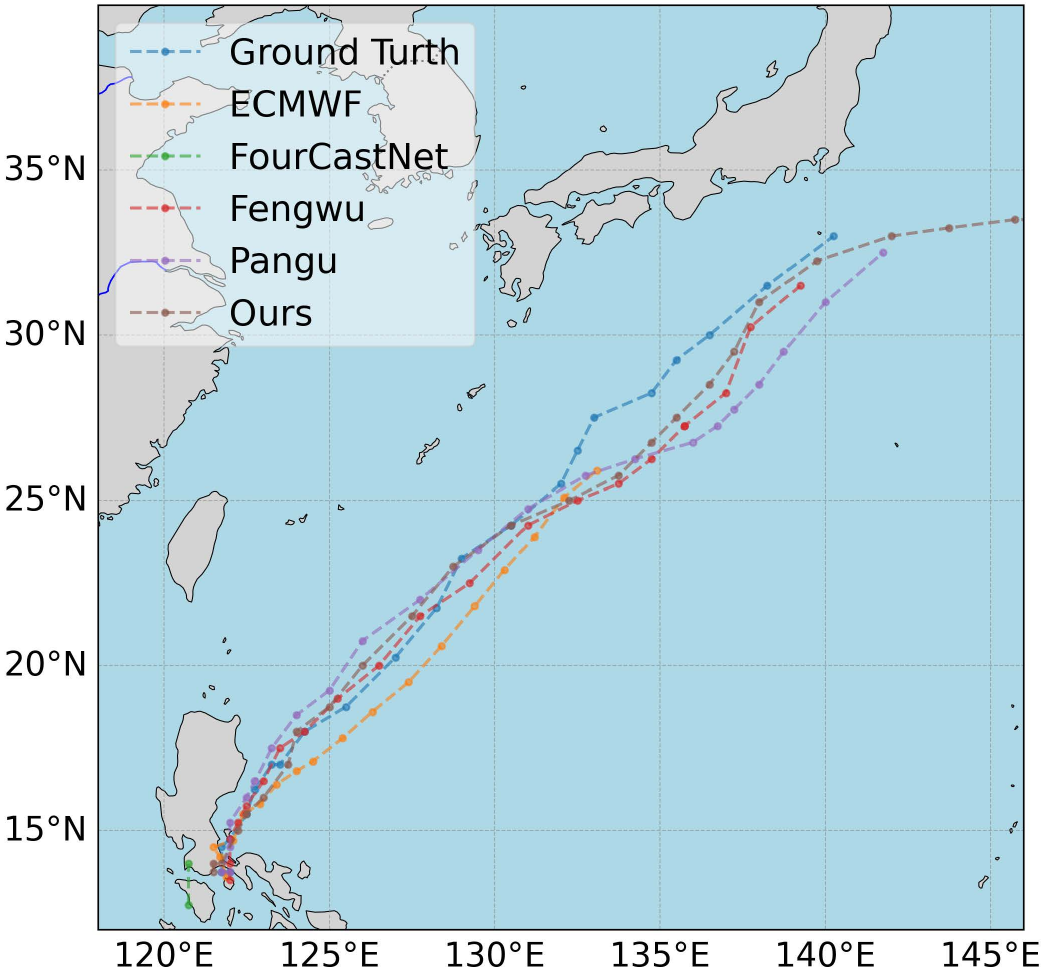}
            \caption{Typhoon Ewiniar (2024.05)}
        \end{subfigure}%
        \hfill
        \begin{subfigure}{0.5\textwidth}
            \centering
            \captionsetup{justification=centering}
            \includegraphics[width=1.0\linewidth]{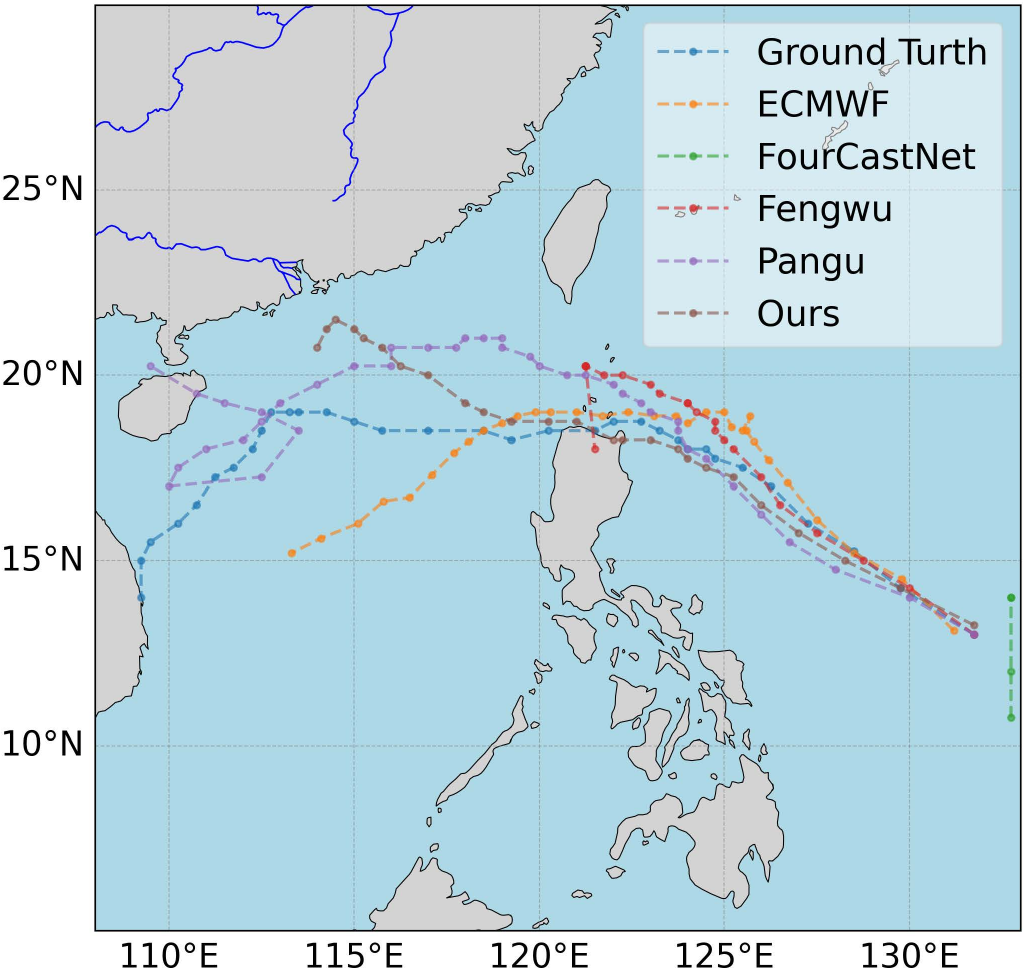}
            \caption{Typhoon Yinxing (2024.11)}
        \end{subfigure}

        \begin{subfigure}{1.0\textwidth}
            \centering
            \captionsetup{justification=centering}
            \includegraphics[width=1.0\linewidth]{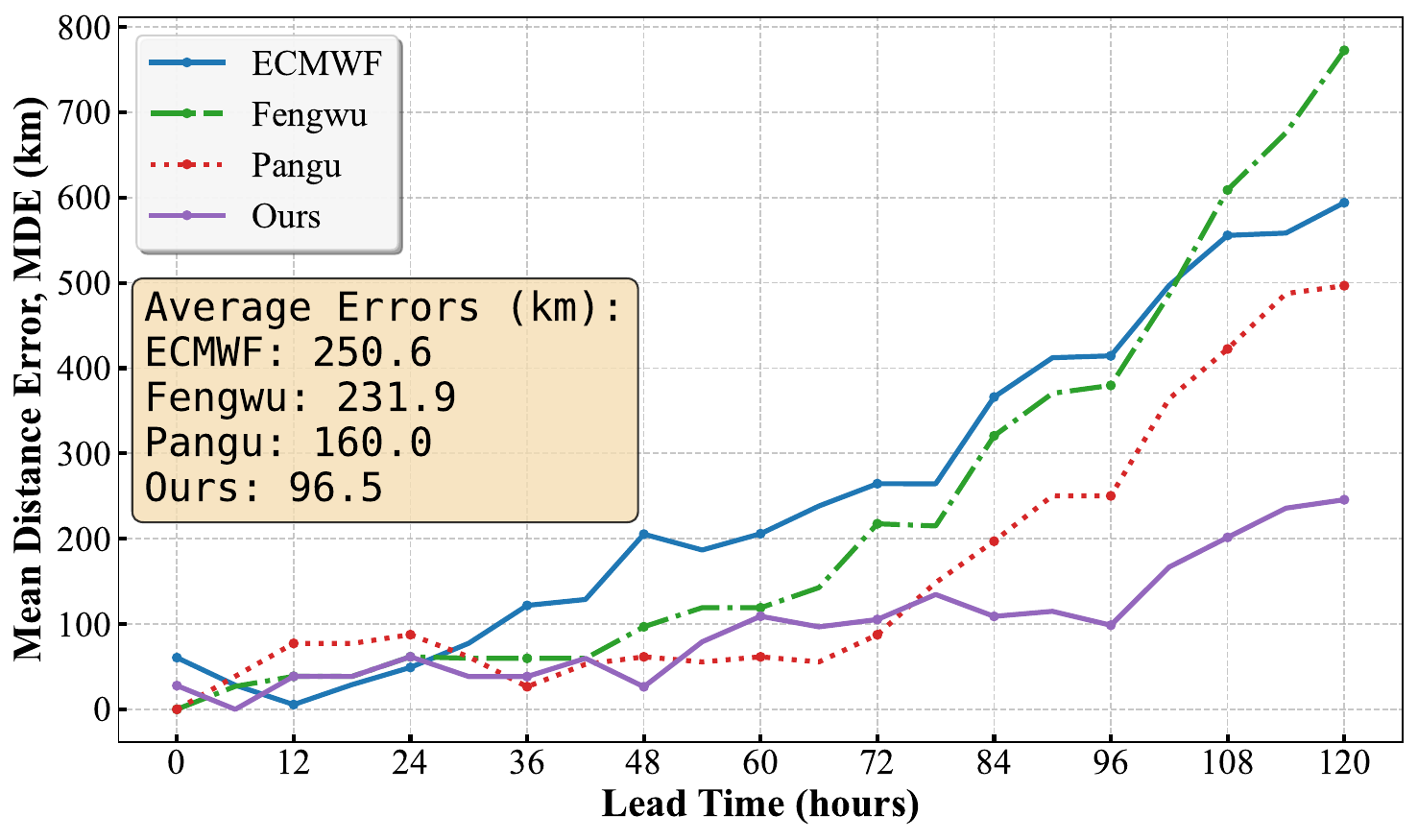}
            \vspace{-0.6cm} 
            \caption{Typhoon Yinxing (2024.11)}
        \end{subfigure}

        \vspace{-3mm}
        \caption{Typhoon Track Assessment. (a) and (b) are the Typhoon Track Comparison in Typhoon Ewiniar and Yinxing, respectively. (c) is 5-day comparative analysis of Mean Distance Error, MDE (km) \boldmath{$\downarrow$} on Yinxing (Ewiniar is provided in Appendix).}
        \vspace{-4mm}
        \label{fig:typhoon}
    \end{minipage}
\end{figure}

\begin{table}[t]
    \small
    \caption{Comparison results of RMSE$\downarrow$ between deterministic forecast and ensemble forecast (ENS), the best are in \textbf{bold}.}
    \vspace{-3mm}
    \centering
    \begin{small}
            \renewcommand{\multirowsetup}{\centering}
            \setlength{\tabcolsep}{3.8pt} 
            \begin{tabular}{l|cccc}
            \toprule
            \multirow{2}{*}{ Model }      & \multicolumn{4}{c}{Forecast Day }   \\ 
            \cmidrule{2-5} 
                              & 7-day                               & 8-day                               & 9-day                               & 10-day                              \\ \midrule
            Pangu             & 0.4875                              & 0.5321                              & 0.5742                              & 0.6213                              \\
            Pangu (ENS)       & 0.4435                              & 0.4743                              & 0.4979                              & 0.5205                              \\
            Graphcast         & 0.4440                              & 0.4923                              & 0.5346                              & 0.5823                              \\
            Graphcast (ENS)   & 0.4412                              & 0.4759                              & 0.5072                              & 0.5331                              \\
            Fuxi              & 0.5928                              & 0.6314                              & 0.6604                              & 0.6968                              \\
            Fuxi (ENS)        & 0.4898                              & 0.5175                              & 0.5353                              & 0.5498                              \\ 
            OneForecast       & 0.4268          & 0.4834       & 0.5313      & 0.5809          \\
            OneForecast (ENS) & 0.4393 & 0.4699 & 0.4951 & 0.5167 \\

            \midrule
            Ours       & \textbf{0.3892} & 0.4285 & \textbf{0.4708} & \textbf{0.5107} \\
            
            Ours (ENS) & 0.3893 & \textbf{0.4284} & 0.4713 & 0.5113 \\

            \bottomrule
            \end{tabular}
                \end{small}
        
        \vspace{-5mm}
        \label{tab:ens}
\end{table}

\subsection{Main Results}

\textbf{Low-resolution Global Forecasts.} We evaluate the performance with two metrics: RMSE and ACC. Due to significant scale variations across atmospheric variables, direct comparison using absolute values is infeasible. We therefore present normalized RMSE and ACC scores in \cref{tab:global}, where STCast demonstrates consistent superiority over baselines across all benchmarks, with particularly significant gains in long-term predictions. Further validation through real-value RMSE and ACC comparisons (1-10 day) in \cref{fig:global} confirms STCast's state-of-the-art performance across multiple variables. This enhancement is attributed to our month-specific training strategy, which effectively captures both seasonal dependencies and month-to-month variations in atmospheric system. Complementary visualization in \cref{fig:vis} compares spatial error distributions for five key variables across three methods, providing qualitative evidence of STCast's reduced prediction uncertainty.


\noindent\textbf{High-resolution Regional Forecasts.} As demonstrated in \cref{fig:sketch}, we compare mean RMSE and ACC scores across five surface variables against Direct Training and OneForecast \cite{oneforecast}. Quantitative analysis reveals that direct-trained STCast (without dynamic boundary) and OneForecast achieve comparable performance. However, implementing our dynamic boundary condition in STCast yields significant improvements: RMSE decreases by 0.05 while ACC increases by 0.1. This enhancement confirms the critical role of adaptive boundary modeling in regional forecasting systems. Complementary visualization of 6-hour regional forecasts for X-direction wind at 10m height (U10) and Mean Sea-Level pressure (MSL) in \cref{fig:region} provides validation. Error analysis demonstrates STCast's superior accuracy, achieving near-zero errors of 0.7\% for U10 and 0.1\% for MSL - substantially lower than competitors.


\noindent\textbf{Extreme Events Assessment.} Extreme weather events, particularly tropical cyclones, pose significant societal risks that demand accurate forecasting capabilities~\cite{wang2025global,wang2025vqlti}. To evaluate STCast's performance under such critical conditions, we analyze two recent typhoon events: Typhoon Ewiniar (May 2024) and Typhoon Yinxing (November 2024)~\cite{ying2014overview, lu2021western}. As visualized in \cref{fig:typhoon}a-b, STCast's 72-hour track forecasts demonstrate substantially closer alignment with observed paths compared to ECMWF~\cite{molteni1996ecmwf}, FourCastNet~\cite{fourcastnet}, Pangu-Weather~\cite{panguweather}, and FengWu~\cite{fengwu} for both systems. Quantitative analysis in \cref{fig:typhoon}c reveals that while STCast achieves comparable performance to others in short-term forecasts, it significantly outperforms them in long-term predictions for Typhoon Yinxing, attaining a mean error of only 96.5 km compared to 160 km for the next-best Pangu-Weather. This improved accuracy, coupled with STCast's consistent capability to capture cyclones' structural development and translational dynamics, highlights its strong potential for extreme event forecasting.

\noindent\textbf{Long-term and Ensemble Weather Forecasting.} As illustrated in \cref{fig:long}, we compare 100-day Z500 predictions from top-performing models. Graphcast exhibits significant predictive degradation in high-latitude regions, while OneForecast maintains functionality but shows substantial poleward deviations from ground truth. In contrast, STCast demonstrates consistent alignment with observations across all latitudes, despite minor localized discrepancies. These long-term forecasts confirm STCast's superior performance in high-latitude prediction tasks. We further evaluate ensemble forecasting capabilities using 10 initial conditions (commencing 00:00 UTC 1 January 2020 at 12-hour intervals) in \cref{tab:ens}. Quantitative assessment via normalized Mean RMSE demonstrates that both STCast and its ensemble variant significantly outperform four competing methods, with our ensemble approach achieving the lowest error distribution across all initialization times.


\noindent\textbf{More Competitors.} ClimaX~\cite{climax}, VAMoE~\cite{vamoe}, EWMoE~\cite{ewmoe}, Keisler~\cite{keisler}, Stormer~\cite{stormer}, FourCastNet~\cite{fourcastnet}, ClimODE~\cite{climode}, WeatherGFT~\cite{weathergft}, and GenCast~\cite{gencast} are all provided in Appendix.

\subsection{Ablation Study}

To further verify the effectiveness of the proposed modules, we conduct comprehensive ablation studies reported in \cref{tab:ablation}. The experiments are split into two groups. \textbf{High-resolution regional forecasts:} (1) STCast w/o SAA: we remove the SAA module and follow the same protocol as OneForecast to predict regional variables; (2) STCast w/o Global-Regional Distribution: we discard the global-regional distribution initialization in SAA and instead use Xavier uniform initialization for the global-regional correlation; (3) STCast w SAA: the complete STCast. \textbf{Low-resolution global forecasts:} (4) STCast w/o TMoE: we replace the Temporal Mixture-of-Experts (TMoE) with MLP block; (5) STCast w/o Month Embedding: we remove the month embedding from TMoE and fall back to a classical Mixture-of-Experts; (6) STCast w TMoE: the complete STCast. Comparisons among (1)–(3) and (4)–(6) reveal that removing any component consistently degrades performance on both regional and global tasks. While the absence of SAA or TMoE causes noticeable drops, the most substantial drops occur when eliminating global-regional distribution (regional: +0.22 RMSE in 10-day) and month embedding (global: +0.13 RMSE in 10-day). These results confirm the critical roles of every component and setting in enhancing the overall effectiveness of STCast.


\textbf{ TMoE. } As illustrated in \cref{tab:ablation-tmoe}, we conduct ablation studies to evaluate the impact of experts' numbers in TMoE: (1) 4 \textbf{w} G: 4 experts with quarterly allocation and Gaussian embedding; (2) 6 \textbf{w} G: 6 experts with two-month allocation and Gaussian embedding; (3) 12 \textbf{w} G: 12 experts with monthly allocation and Gaussian embedding; (4) 12 \textbf{w/o} G: 12 experts with monthly allocation and one-hot embedding. From settings (1) to (3), results show that increasing the number of experts improves performance, albeit with a significant rise in the parameters. Further increasing the experts' number to 24, corresponding to a biweekly allocation scheme, results in the parameter of 1164.7M, which becomes computationally prohibitive. While finer temporal divisions yield better results, a balance between accuracy and efficiency is achieved by setting the experts' number to 12. Comparing (3) and (4), we observe that Gaussian embeddings outperform one-hot embeddings in all settings. Unlike one-hot, which activate only one expert at a time, Gaussian enable the model to capture relationships both within and across months, resulting in better performance.


\textbf{ SAA. }  To evaluate the impact of distance metrics on the SAA, we performed an ablation study comparing three measures in \cref{tab:ablation-saa}: Manhattan, Euclidean, and Great Circle distances. The results show that while the choice of distance metric has a limited effect on overall performance, the Great Circle distance yields the lowest RMSE due to its superior accuracy in measuring distances on the Earth's surface. The learned spatial distribution is illustrated in \cref{fig:saa}.




\begin{table}[t]
    
    \centering
    \caption{ Ablation Studies on TMoE with normalized mean RMSE. `Experts \#' and `G' denote experts' numbers and Gaussian Distribution in TMoE, respectively. }
    \vspace{-3mm}
    \begin{small}
            \renewcommand{\multirowsetup}{\centering}
            \setlength{\tabcolsep}{2.8pt} 
            \begin{tabular}{l|ccccc|c}
                \toprule
                \multicolumn{7}{c}{ \textbf{Low-resolution Global Forecasts } (RMSE$\downarrow$) } \\
                
                \midrule
                Experts \# & \multicolumn{1}{c}{6-hour} & \multicolumn{1}{c}{1-day} & \multicolumn{1}{c}{4-day} & \multicolumn{1}{c}{7-day} & \multicolumn{1}{c}{10-day} & Params  \\     
                \midrule

                4 \pmb{w} G & 0.0750  & 0.1394  & 0.3147  & 0.5066  & 0.6433 & 484.9M  \\

                6 \pmb{w} G & 0.0738  & 0.1329  & 0.2808  & 0.4555     & 0.5932 & 541.5M \\
                
                12 \pmb{w} G & \textbf{0.0617}  & \textbf{0.1197}  & \textbf{0.2578}  & \textbf{0.4348}  & \textbf{0.5763} & 654.8M \\

                \midrule

                12 \pmb{w/o} G & 0.0734  & 0.1351  &0.2795  &  0.4509  &  0.5911  & 654.8M \\
                
                \bottomrule
            \end{tabular}
	\end{small}
    \vspace{-2mm}
    \label{tab:ablation-tmoe}
\end{table}

\begin{table}[t]
    
    \small
    \centering
    \caption{ Ablation Studies on SAA with normalized mean RMSE. }
    \vspace{-3mm}
    \begin{small}
            \renewcommand{\multirowsetup}{\centering}
            \setlength{\tabcolsep}{2.8pt} 
            \begin{tabular}{l|ccccc}
                \toprule
                \multicolumn{6}{c}{ \textbf{High-resolution Regional Forecasts} (RMSE$\downarrow$) } \\
                
                \midrule
                Distance Metric  &   \multicolumn{1}{c}{6-hour} & \multicolumn{1}{c}{1-day} & \multicolumn{1}{c}{4-day} & \multicolumn{1}{c}{7-day} & \multicolumn{1}{c}{10-day}   \\   
                \midrule

                
                Manhattan  & 0.0493  & 0.0854  & 0.2068  & 0.3712  & 0.4945  \\
                
                Euclidean & 0.0501  & 0.0862  & 0.2093  & 0.3892  & 0.5088  \\

                Great Circle & 0.0478  & 0.0845  & 0.2051  & 0.3699  & 0.4921   \\

                \bottomrule
            \end{tabular}
	\end{small}
    \vspace{-4mm}
    \label{tab:ablation-saa}
\end{table}

\section{Conclusion}
\label{sec:conclusion}



In this work, we introduce an adaptive attention map within the Spatial-Aligned Attention (SAA) module to provide dynamic boundary conditions for regional forecasting. Beyond regional task, we embed a Temporal Mixture-of-Experts (TMoE) into the Spatial-Temporal Forecasting (STCast), casting weather prediction as a multi-task problem and delegating monthly sub-tasks to specialized experts. Consequently, STCast simultaneously addresses 4 distinct challenges: low-resolution global forecasting, high-resolution regional forecasting, extreme-event assessment, and ensemble weather forecasting. Experiments and ablation studies confirm that STCast consistently outperforms competing methods across all evaluated scenarios.

\section{Acknowledgements}

This research was supported by fundings from the Hong Kong RGC General Research Fund (152228/23E, 162161/24E, 162116/25E, 162180/25E), National Natural Science Foundation of China (NSFC) Key Program (No.62532005), Collaborative Research Fund (No. C1042-23GF, No. 5097-25G), NSFC/RGC Collaborative Research Scheme (Grant No. 62461160332 \& CRS\_HKUST602/24), Research Impact Fund (No. R5011-23F), Areas of Excellence Scheme (AoE/E-601/22-R), and the InnoHK (HKGAI).

{
    \small
    \bibliographystyle{ieeenat_fullname}
    \bibliography{main}
}


\end{document}